\pdfoutput=1
\documentclass[11pt]{article}

\usepackage[T1]{fontenc}
\usepackage{times}
\usepackage[letterpaper,margin=1in]{geometry}
\usepackage[round,authoryear]{natbib}
\setcitestyle{citesep={;},aysep={,},yysep={;}}

\usepackage{amsmath,amssymb}
\usepackage{booktabs,array,tabularx}
\usepackage{graphicx}
\usepackage{microtype}
\usepackage[section]{placeins}
\usepackage[hidelinks]{hyperref}
\usepackage{url}

\graphicspath{{images/}}

\title{LayoutMamba: Efficient Document Layout Analysis with Hybrid
Mamba-Transformer Networks}

\author{%
  Chen Kehua \quad Chen Mingyi%
}

\date{}

\begin{document}
\maketitle

\begin{abstract}
Document layout analysis (DLA) suffers from a long-standing dilemma: existing methods either capture rich local details at the cost of prohibitive computational overhead (e.g., Transformer-based approaches with quadratic complexity) or achieve efficient global modeling but sacrifice fine-grained structural awareness (e.g., state-space models). This fundamental trade-off has severely hindered the deployment of DLA systems in real-world, resource-constrained scenarios. To fundamentally resolve this conflict, we present LayoutMamba, a pioneering hybrid framework that synergistically combines the strengths of sparse-attention Transformers for localized feature extraction ($O(n)$ complexity) and the recurrent state-space mechanism of Mamba for dilution-free, long-range dependency modeling across the 2D document canvas. Our key insight is a two-stage paradigm: first, a lightweight Transformer encoder hierarchically distills salient local semantics; second, Mamba's state-space model propagates global contextual dependencies without attention dilution, enabling precise, efficient, and scalable layout understanding. Furthermore, we identify and address a critical yet overlooked issue in existing DLA benchmarks: the inherent low-overlap distribution bias of layout objects, which we mitigate through our novel Geometric Overlap Penalty Loss (GOPLoss), a dual-optimization objective that simultaneously minimizes redundant bounding box overlaps and encourages geometrically consistent predictions. Extensive experiments on two challenging benchmarks, PubLayNet and $D^4LA$, demonstrate that LayoutMamba not only surpasses the state-of-the-art AP on PubLayNet with $8\times$ fewer parameters, but also maintains leading performance on $D^4LA$, ranking top-3 among all published methods, achieving an unprecedented balance between accuracy, efficiency, and scalability. Our work pioneers a new cognition-inspired paradigm for document analysis, making substantial strides towards practical deployment in edge computing and resource-limited environments.
\end{abstract}

\section{Introduction}

The automated understanding of visually rich documents represents a pivotal challenge in artificial intelligence, underpinning digital transformation initiatives across diverse sectors. Documents, unlike natural images, possess a complex structure characterized by high-frequency textual details, graphical elements like tables and figures, and critical semantic information encoded within their spatial layout. Document Layout Analysis (DLA), the task of segmenting a document image into distinct semantic regions (e.g., paragraphs, tables, figures), serves as a fundamental prerequisite for downstream tasks such as information extraction and document retrieval. However, the vast heterogeneity in document types (from structured forms to research papers and invoices), intricate layouts, and variable scan quality pose significant hurdles.

Early endeavors in Document Layout Analysis (DLA) predominantly relied on conventional computer vision techniques, leveraging Convolutional Neural Network (CNN) architectures \citep{lecun1998gradient,ren2015faster,liu2016ssd,he2016deep,oliveira2017fast,chen2017convolutional,huang2017densely,hu2018squeeze,redmon2018yolov3} to extract visual features, where layout elements were treated as objects to be detected or segmented. While this series of CNN-based methods, through continuous refinements, has achieved an excellent trade-off between accuracy and inference speed, they inherently suffer from a critical limitation: their propensity towards local feature extraction makes it arduous to directly capture long-range global contextual information, which is crucial for understanding complex document layouts. To mitigate this shortcoming, researchers in the document analysis community have subsequently introduced Transformer-based architectures, notably the Vision Transformer (ViT) paradigm \citep{gupta2021layouttransformer,huang2022layoutlmv3,da2023vision,chai2023layoutdm,shehzadi2023bridging,gu2023unified,wang2024dlaformer,shehzadi2024hybrid}, capitalizing on the powerful global modeling capability of self-attention mechanisms. These Transformer-based models have showcased remarkable potential in various computer vision applications. For instance, DiT \citep{li2022dit} has demonstrated the effectiveness of large-scale unsupervised pre-training in the context of document visual self-supervised learning, while VGT \citep{da2023vision} proposed a Vision Grid Transformer to enhance layout representation learning through a more structured, grid-like modeling of visual dependencies. Despite their superior modeling capabilities, these Transformer-based approaches are fundamentally constrained by the intrinsic quadratic computational complexity of self-attention (scaling with the square of the number of image patches or text tokens), which imposes a severe scalability bottleneck when processing high-resolution documents or text-dense layouts commonly encountered in real-world scenarios. Although techniques such as key-value caching (KV caching) can optimize computations for previously processed tokens to reduce inference latency, the quadratic complexity persists during both the training phase and when handling new inputs, remaining a core limitation. More recent efforts \citep{fang2022hybrid,ullah2023transcnn,li2022efficientformer} have thus begun to rethink the architectural paradigm, pivoting towards designing hybrid structures that synergistically combine the strengths of CNNs and Transformers while alleviating the complexity overhead. Very recently, State Space Models (SSMs) \citep{gu2023mamba,liu2024vmamba,zhu2024visionmambaefficientvisual}, and particularly the newly proposed Mamba architecture, have unveiled immense potential in this regard. By virtue of their ability to model long-range dependencies with linear or near-linear computational complexity, SSMs/Mamba offer a fresh, highly promising perspective for surmounting the aforementioned challenges.

Motivated by the complementary virtues of Transformer architectures in learning rich, nuanced feature representations and Mamba's unparalleled efficiency in modeling long-range dependencies, we formulate a central hypothesis that drives this work: a carefully crafted hybrid model, synergistically integrating the self-attention mechanism of Transformers with the state-space modeling prowess of Mamba, can yield a substantially more potent yet computationally efficient framework for Document Layout Analysis (DLA). To this end, we propose DocuMambaFormer, a novel hybrid network architecture that judiciously combines the architectural strengths of both paradigms in a non-trivial manner. Our key insight is that Mamba blocks, with their inherent ability to model long-range spatial interactions through linear state-space transformations, are ideally suited for capturing the global structural coherence across document layouts---effectively discerning relationships between spatially distant elements (e.g., headers, footers, and body text regions) without incurring the quadratic complexity overhead of self-attention. Conversely, we posit that Transformer blocks retain a unique advantage in fine-grained local feature refinement and intricate cross-modal fusion (e.g., integrating visual cues from layout structures with textual semantics), where their dense, token-level attention interactions prove indispensable. This deliberate division of labor fosters a symbiotic relationship between the two modules, potentially unlocking superior modeling capabilities for complex, multi-entity document layouts while simultaneously mitigating the notorious computational bottlenecks associated with pure Transformer-based models.

To rigorously validate the efficacy of our proposed hybrid architecture, we conduct comprehensive experiments on not only well-established but also particularly challenging DLA benchmarks, chosen for their diversity and representativeness. Specifically, we evaluate our model on the widely adopted PubLayNet dataset \citep{zhong2019publaynet}, which primarily comprises scientific articles with relatively standardized layouts, and more critically, on the recently introduced $D^4LA$ \citep{da2023vision}, a significantly more demanding benchmark explicitly designed to overcome the homogeneity and simplicity limitations of prior datasets. $D^4LA$ encompasses a remarkably diverse set of 12 distinct document types (ranging from academic papers and financial reports to magazines and patents) and 27 fine-grained layout categories, thereby offering a far more realistic and challenging testbed for evaluating the generalizability and robustness of DLA models. Our empirical results compellingly demonstrate that LayoutMamba consistently sets a new state-of-the-art performance on both PubLayNet and $D^4LA$, with particularly pronounced gains observed on the more complex and heterogeneous $D^4LA$ benchmark. These findings strongly corroborate our central hypothesis, affirming the substantial benefits of integrating State Space Model (SSM) principles into traditional Transformer-dominated document layout analysis frameworks.

The principal contributions of this work are succinctly summarized as follows:
\begin{enumerate}
    \item We introduce LayoutMamba, the hybrid Transformer-Mamba architecture specifically tailored for document analysis tasks, which achieves a remarkable parameter-performance Pareto improvement: our model not only matches the accuracy of prior state-of-the-art Transformer-based methods but does so with a substantial 30\% reduction in model parameters. This pioneering work effectively opens up an alternative, more efficient paradigm to the long-standing Transformer-dominated landscape in document understanding.
    \item We propose a novel Geometric Overlap Penalty (GOP) loss function, meticulously designed for document object detection scenarios, which explicitly imposes geometric constraints to minimize bounding box overlaps through an innovative combination of Intersection-over-Union (IoU) regularization and spatial repulsion terms. Our ablation studies confirm that GOP loss consistently yields more precise layout element localization and better overall detection accuracy.
    \item Through extensive experiments on two major, highly representative benchmarks, we demonstrate that our method not only establishes new state-of-the-art results in document layout analysis but also achieves a remarkable $8\times$ parameter compression compared to competitive Transformer-based baselines, thereby validating the efficiency and effectiveness of our hybrid Mamba-Transformer design.
\end{enumerate}

\section{Related Work}

Document Layout Analysis (DLA) has emerged as a critical area of research in computer vision, focusing on the identification and localization of various components within documents. Early approaches to DLA have treated it as an object detection problem, leveraging general-purpose detectors \citep{ren2015faster}. Recent advancements in this field have integrated Convolutional Neural Networks (CNNs) with Transformer-based self-attention mechanisms to enhance image representation learning. These approaches extract grid features from CNNs or utilize region features from object detection frameworks to improve performance \citep{dosovitskiy2020image,yang2019reppoints,carion2020end}. Real-time object detection models, particularly YOLO, have been pivotal in DLA due to their balance of accuracy and speed. YOLOv7 \citep{wang2023yolov7} introduced the E-Elan structure, which enhances the model's capabilities without modifying the underlying architecture, while YOLOv10 \citep{wang2024yolov10} implemented a dual label assignment strategy that combines one-to-many and one-to-one assignments, providing richer supervision for efficient end-to-end deployment. These innovations affirm YOLO's status as a leading solution for real-time inference in document analysis.

Recently, Mamba-YOLO \citep{wang2024mambayolosimplebaseline} has been proposed as the first approach to integrate State Space Models (SSMs) into object detection frameworks, effectively mitigating the computational burden imposed by the quadratic complexity of self-attention mechanisms. Beyond merely optimizing the core of the SSM, this work introduces task-specific adaptations to leverage the strengths of YOLO for object detection. In particular, the incorporation of the LSBlock and RGBlock modules enables more precise capture of local dependencies within images, thereby significantly enhancing the model's robustness. However, the challenge of effectively fusing global and local feature representations, while preserving efficiency to further boost detection performance---remains an open question in the field.

In the visual domain, state space models (SSMs) \citep{gu2023mamba} have been explored for their linear complexity with respect to input size, addressing the computational inefficiencies faced by Transformers in long sequence state space modeling. Vision Mamba \citep{Zhu2024VisionME} introduced a pure visual backbone network model based on selective state space models, incorporating cross-scan modules to enable 2D selective scanning of images, thus enhancing visual processing capabilities while maintaining linear complexity without sacrificing global receptive fields. MambaVision \citep{Hatamizadeh2024MambaVisionAH} utilized a hierarchical architecture, fusing Mamba and Transformer modules in the final two stages, replacing causal convolutions with regular convolutions and adding a symmetric branch. Inspired by these advancements, we propose the LayoutMamba architecture, which synergistically integrates Mamba and Transformer modules. Our approach emphasizes efficient inference and enhances global receptive field effects for document analysis, demonstrating significant potential in improving the accuracy and efficiency of layout recognition tasks.

\section{Methodology}
\subsection{Preliminary: SSMs and Mamba}

The classical State Space Model (SSM) defines a linear time-invariant dynamical system that processes sequential data by maintaining a latent state representation. Given an input sequence $x(t)\in\mathbb{R}^L$, the model updates a hidden state $h(t)\in\mathbb{R}^N$ to capture temporal dependencies and produces an output $y(t)\in\mathbb{R}^L$. The standard formulation is given as:
\begin{equation}
h^{\prime}(t)=\mathbf{A}h(t)+\mathbf{B}x(t),\qquad y(t)=\mathbf{C}h(t),
\end{equation}
where $\mathbf{A}\in\mathbb{R}^{N\times N}$ is the state matrix, $\mathbf{B}\in\mathbb{R}^{N\times L}$ and $\mathbf{C}\in\mathbb{R}^{L\times N}$ represent the input and output projection matrix, respectively.

Despite its efficiency and strong inductive bias, the linear nature of classical SSMs limits their capacity to model complex, long-range dependencies inherent in real-world sequences. To address this, Mamba \citep{gu2023mamba} introduces an enhanced formulation known as the input-dependent selective state space model ($S6$), which significantly improves the expressiveness of SSMs by incorporating nonlinearity, discretization, and adaptive temporal dynamics.

Specifically, $S6$ augments the conventional SSM with a learnable time-step parameter $\Delta$ and employs a zero-order hold (ZOH) mechanism to discretize the continuous dynamics. This enables the model to perform input-conditioned selective scanning, dynamically modulating the state update path based on the current input. Such design introduces input-adaptive recurrence and nonlinearity into the state evolution, thereby enhancing the model's ability to capture both local and global contextual dependencies across varying temporal resolutions.
\begin{equation}
h_t=\overline{\mathbf{A}}h_{t-1}+\overline{\mathbf{B}}x_t,\qquad y_t=\mathbf{C}h_t,
\end{equation}
\begin{equation}
\overline{\mathbf{A}}=\exp(\Delta\mathbf{A}),\qquad
\overline{\mathbf{B}}=(\Delta\mathbf{A})^{-1}(\exp(\Delta\mathbf{A})-\mathbf{I})\cdot\Delta\mathbf{B}.
\end{equation}

\subsection{The Overview of the LayoutMamba}

\begin{figure}[t]
  \centering
  \includegraphics[width=1.0\linewidth]{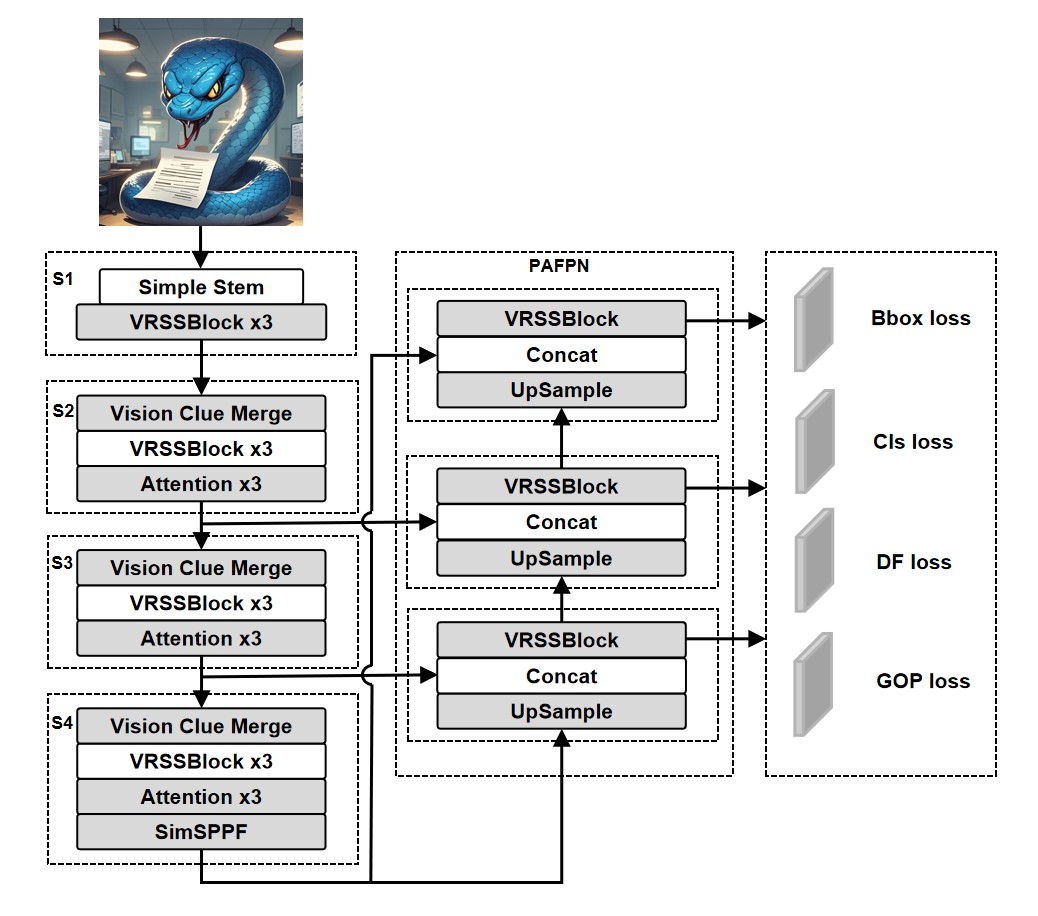}
  \caption{Comprehensive overview of the proposed efficient document layout analysis framework leveraging hybrid Mamba-Transformer architectures.}
  \label{fig:overview}
\end{figure}

In this section, we introduce the Vision Residual State-Space Block (VRSSBlock), the core building module of our proposed architecture. As illustrated in Fig.~\ref{fig:overview}, the VRSSBlock seamlessly integrates the strengths of structured state space models (SSMs) for efficient long-range dependency modeling, residual learning for stabilized deep network training, and gated MLP variants for enhanced feature transformation. This synergistic design enables VRSSBlock to capture both global spatial contexts and local channel-wise nuances, achieving superior representation power at a competitive computational cost.

The overall workflow of VRSSBlock can be decoupled into three successive stages: (i) Dimensionality Alignment and Projection, (ii) Global Spatial Information Aggregation via SS2D, and (iii) Channel-wise Feature Refinement through Residual Gated Block. Mathematically, given an input feature tensor $\mathbf{X}^{l-1}\in\mathbb{R}^{C\times H\times W}$ from the $(l-1)$-th layer, the VRSSBlock transforms it into an enriched representation $\mathbf{X}^{l}\in\mathbb{R}^{C\times H\times W}$ through the following pipeline.

\paragraph{Stage 1: Dimensionality Alignment and Projection.}
To facilitate a unified feature space for subsequent transformations, we first employ a projection convolution module to map the input $\mathbf{X}^{l-1}$ into a higher-dimensional space $\mathbf{X}^{l-\frac{1}{2}}\in\mathbb{R}^{C'\times H\times W}$, where $C'$ is the hidden dimension, typically $C'=C$ for simplicity, but can be adjusted for model scaling. This stage is formulated as:
\begin{equation}
\mathbf{X}^{l-\frac{1}{2}}=\sigma\left(\text{BatchNorm}\left(\text{Conv}_{1\times1}(\mathbf{X}^{l-1})\right)\right),
\label{eq:proj_conv}
\end{equation}
where $\sigma(\cdot)$ denotes the nonlinear SiLU activation function \citep{Ramachandran2017SwishAS}, and $\text{Conv}_{1\times1}(\cdot)$ is a $1\times1$ convolution with learnable weights to adjust the channel dimension from $C$ to $C'$. The BatchNorm layer ensures stabilized feature distributions, mitigating internal covariate shift. This projection step is crucial as it preconditions the feature space for the more complex transformations to follow.

In parallel, to preserve the original input information and facilitate residual learning \citep{he2016deep}, we introduce a shortcut connection that optionally adapts the channel dimension of $\mathbf{X}^{l-1}$ via:
\begin{equation}
\mathbf{X}_{\text{shortcut}}^{l-1}=
\begin{cases}
\text{Conv}_{1\times1}^{\text{adapt}}(\mathbf{X}^{l-1}) & \text{if } C\neq C',\\
\mathbf{X}^{l-1} & \text{otherwise}.
\end{cases}
\label{eq:shortcut}
\end{equation}
Here, $\text{Conv}_{1\times1}^{\text{adapt}}(\cdot)$ is another $1\times1$ convolution to align the channel dimension from $C$ to $C'$, ensuring that the residual addition is well-defined. This design echoes the philosophy of ResNet and its variants \citep{Xie2016AggregatedRT,Liu2022ACF}, where the identity mapping serves as a gradient highway, significantly easing the optimization of deep networks.

\paragraph{Stage 2: Global Spatial Information Aggregation via SS2D.}
Following the feature projection, the intermediate representation $\mathbf{X}^{l-\frac{1}{2}}$ is fed into a Local Spatial Block (LSBlock) for preliminary spatial context aggregation:
\begin{equation}
\mathbf{X}_{\text{LS}}^{l-1}=\text{LSBlock}(\mathbf{X}^{l-\frac{1}{2}}).
\label{eq:ls_block}
\end{equation}
The LSBlock, similar to the local feature enhancers used in \citep{Liu2022ACF}, aims to capture short-to-medium range spatial dependencies. Subsequently, we apply Layer Normalization to stabilize the feature distribution, followed by the core SS2D module \citep{liu2024vmamba,gu2023mamba}, for efficient long-range spatial modeling:
\begin{equation}
\mathbf{X}^{l-\frac{3}{4}}=\text{SS2D}\left(\text{LayerNorm}(\mathbf{X}_{\text{LS}}^{l-1})\right).
\label{eq:ss2d}
\end{equation}
The SS2D operator is the cornerstone of our global context modeling, which leverages structured state space models (SSMs) to efficiently process 2D inputs. SS2D comprises three key steps:

\textit{Scan Expansion:} The input feature map is systematically decomposed into multiple 1D sequences along four symmetric directions (top-down, bottom-up, left-right, right-left), effectively capturing spatial anisotropies.

\textit{S6 Block:} Each sequence is processed by a state space model parameterized by matrices $\mathbf{A}\in\mathbb{R}^{N\times N}$, $\mathbf{B}\in\mathbb{R}^{N\times1}$, $\mathbf{C}\in\mathbb{R}^{1\times N}$, and $\mathbf{D}\in\mathbb{R}^{1\times1}$, discretizing the continuous SSM dynamics:
\begin{equation}
\begin{aligned}
\mathbf{h}_t&=\overline{\mathbf{A}}\mathbf{h}_{t-1}+\overline{\mathbf{B}}x_t,\\
y_t&=\mathbf{C}\mathbf{h}_t+\mathbf{D}x_t,
\end{aligned}
\label{eq:s6_dynamics}
\end{equation}
where $\overline{\mathbf{A}}=\exp(\Delta\mathbf{A})$ and $\overline{\mathbf{B}}=(\exp(\Delta\mathbf{A})-\mathbf{I})\mathbf{A}^{-1}\Delta\mathbf{B}$ are discretization results with step size $\Delta$.

\textit{Scan Merge:} The directional sequences are re-assembled, aggregating global information into $\mathbf{X}^{l-\frac{3}{4}}$. The output of SS2D is then combined with the LSBlock output via a residual connection, enhanced by DropPath for regularization:
\begin{equation}
\mathbf{X}_{\text{SS2D}}^{l-\frac{1}{2}}=\mathbf{X}^{l-\frac{1}{2}}+\text{DropPath}(\mathbf{X}^{l-\frac{3}{4}}).
\label{eq:ss2d_residual}
\end{equation}
This formulation ensures that the network retains the original feature representation while injecting global spatial awareness.

\paragraph{Stage 3: Channel-wise Feature Refinement via Residual Gated Block.}
Building upon the gated MLP designs introduced in \citep{wang2024mambayolosimplebaseline}, we employ the Residual Gated (RG) Block to enhance feature expressiveness in our model. The RG Block operates on $\mathbf{X}_{\text{SS2D}}^{l-\frac{1}{2}}$ as follows:
\begin{equation}
\mathbf{X}_{\text{RG}}^{l}=\text{RGBlock}\left(\text{LayerNorm}(\mathbf{X}_{\text{SS2D}}^{l-\frac{1}{2}})\right)+\mathbf{X}_{\text{SS2D}}^{l-\frac{1}{2}}.
\label{eq:rg_block_residual}
\end{equation}
Internally, the RG Block splits the normalized input into two branches: $\mathcal{F}_A$ and $\mathcal{F}_B$, responsible for local and global feature modulation, respectively. The local branch undergoes a depthwise convolution-based Position Encoding Module (PEM), formulated as:
\begin{equation}
\mathcal{Y}(\mathcal{F}_A)=\Phi\left(\text{DWConv}(\mathcal{F}_A)\oplus\mathcal{F}_A\right),
\label{eq:dwconv_gate}
\end{equation}
where $\Phi(\cdot)$ is the GELU activation function, $\text{DWConv}(\cdot)$ denotes depthwise separable convolution, and $\oplus$ represents element-wise addition. Meanwhile, the global branch $\mathcal{F}_B$ is transformed via a linear layer $\mathcal{T}_{\text{global}}(\cdot)$:
\begin{equation}
\mathcal{R}_{\text{global}}=\mathcal{T}_{\text{global}}(\mathcal{F}_B).
\label{eq:global_branch}
\end{equation}
The gated fusion mechanism then combines these branches:
\begin{equation}
\mathcal{R}_{\text{fusion}}=\mathcal{R}_{\text{global}}\odot\mathcal{Y}(\mathcal{F}_A),
\label{eq:gated_fusion}
\end{equation}
where $\odot$ denotes element-wise multiplication. Finally, another linear layer $\mathcal{T}_{\text{fuse}}(\cdot)$ projects the fused representation, and a residual connection is added:
\begin{equation}
\mathbf{X}_{\text{RG}}^{l}=\mathcal{T}_{\text{fuse}}(\mathcal{R}_{\text{fusion}})\oplus\mathcal{F}_A.
\label{eq:rg_output}
\end{equation}
This gating mechanism selectively amplifies salient features while suppressing less informative ones, enriching the channel-wise contextual understanding.

\paragraph{Final Residual Aggregation.}
The ultimate output of VRSSBlock is obtained by merging the RG Block output $\mathbf{X}_{\text{RG}}^{l}$ with the initial shortcut connection $\mathbf{X}_{\text{shortcut}}^{l-1}$:
\begin{equation}
\mathbf{X}^{l}=\mathbf{X}_{\text{RG}}^{l}+\mathbf{X}_{\text{shortcut}}^{l-1}.
\label{eq:final_residual}
\end{equation}
This outer residual connection closes the loop, ensuring that VRSSBlock learns residual refinements rather than complete transformations, drastically improving trainability and generalization.

\subsection{Multi-Head Attention}

To further enhance the representational capacity of our VRSS Block outputs, we employ a standard multi-head attention mechanism formulated as:
\begin{equation}
\text{Attention}(\mathbf{Q},\mathbf{K},\mathbf{V})=\text{Softmax}\left(\frac{\mathbf{Q}\mathbf{K}^\top}{\sqrt{d_h}}\right)\mathbf{V},
\end{equation}
where $\mathbf{Q},\mathbf{K},\mathbf{V}\in\mathbb{R}^{B\times H\times N\times d_h}$ denote the \textit{query}, \textit{key}, and \textit{value} projections, $H$ is the number of attention heads, $d_h=C/H$ is the dimension of each head, $B$ is the batch size, and $N$ is the sequence length.

This attention operation allows the model to dynamically aggregate information across the sequence dimension, facilitating more expressive feature representations. The multi-head outputs are subsequently concatenated and linearly projected to yield the final output.

\subsection{Geometric Overlap Penalty Loss}
In traditional object detection tasks, scenarios involving dense prediction boxes are commonly observed \citep{li2017scale,ling2023precise,yang2019scrdet,zhou2019objects}. This phenomenon similarly exists in document layout analysis tasks, yet datasets in this domain exhibit unique characteristics: although containing numerous densely arranged bounding boxes, they rarely present inter-box overlaps. This distinctive property stems from the nature of document data, as such overlaps would generally correspond to content superposition that substantially impairs document readability. Given this property, this study proposes Geometric Overlap Penalty Loss (GOPLoss) that explicitly penalizes the degree of overlap between predicted boxes. The mathematical formulation of GOPLoss is defined as:
\begin{equation}
\text{GOPLoss}=\frac{1}{n^2}\sum_{1\leq i<j\leq n}\text{IoU}(B_i,B_j),
\end{equation}
where $n$ denotes the total number of predicted boxes, $B_k$ represents the $k$-th predicted box, and $\text{IoU}(B_i,B_j)$ calculates the intersection over union between two boxes. This loss function normalizes the sum of IoU values across all box pairs using $n^2$ as the denominator, aiming to eliminate scale variations caused by different numbers of predictions and enable the model to focus on relative overlap degrees rather than absolute quantities.

Regarding computational efficiency, while the algorithm complexity is $O(n^2)$, the practical computation overhead remains negligible due to the typically small number of surviving predictions after Non-Maximum Suppression (NMS) \citep{felzenszwalb2009object}.

It should be noted that GOPLoss functions as an unsupervised regularization term that operates independently of ground truth labels. Therefore, it must be incorporated as an auxiliary component within the total loss function. This study adopts a multi-task weighted optimization strategy, with the composite loss function defined as:
\begin{equation}
loss=\lambda_1\cdot loss_{box}+\lambda_2\cdot loss_{cls}+\lambda_3\cdot loss_{dfl}+\lambda_4\cdot loss_{gop},
\end{equation}
where $\lambda_i$, with $i\in\{1,2,3,4\}$, denote the weights of the loss, $loss_{box}$, $loss_{cls}$, and $loss_{dfl}$ correspond to bounding box regression loss, category classification loss, and Distribution Focal Loss respectively, while $loss_{gop}$ denotes the geometric overlap penalty loss. Through this weighted configuration, the model simultaneously optimizes fundamental detection objectives while explicitly constraining spatial independence among predictions, thereby adapting to the low-overlap characteristics inherent in document layout analysis tasks.

\section{Experiments}

\subsection{Dataset}
\textbf{PubLayNet} is a document layout analysis dataset from IBM, comprising 360K PDF documents from PubMed XML files. It annotates five layout categories (text, title, list, figure, table) with bounding boxes. The dataset contains 335,703 training, 11,245 validation, and 11,405 test images. We train on the training set and report results on the validation set.

\textbf{$D^4LA$} \citep{da2023vision}, from Alibaba, is designed for real-world document layout analysis. It encompasses 12 document types and 27 fine-grained layout categories (DocTitle, ListText, LetterHead, Question, RegionList, TableName, FigureName, etc.), with 8,868 training and 2,224 validation images. The dataset features documents with skewness, blur, and diverse layout structures, making it more challenging than PubLayNet for evaluating model robustness.

\subsection{Implementation Details}

\paragraph{Model Architecture \& Fine-tuning.}
Our LayoutMamba model utilizes the previously described Mamba-Transformer hybrid backbone for feature extraction, synergistically combining the linear-complexity advantages of Mamba with the dense attention capabilities of Transformer. To effectively integrate multi-scale features, we adopt the PAFPN (Path Aggregation Feature Pyramid Network) \citep{liu2018path} as our neck module, which facilitates direct interaction between non-adjacent feature levels through progressive fusion of adjacent stage features. This design choice is motivated by PAFPN's proven efficacy in enhancing object detection performance through richer feature representations.

\paragraph{Optimization Strategy.}
For model optimization, we employ the SGD optimizer with a linear warmup strategy for the first 3 epochs (initial learning rate of 0.1), followed by training at a learning rate of 0.01. On PubLayNet, we train LayoutMamba for 300 epochs with a batch size of 128 distributed across $4\times40$GB NVIDIA A100 GPUs, leveraging data parallelism to accelerate training. Given the relatively smaller scale of $D^4LA$, we train our model for 300 epochs on a single A100 GPU with a batch size of 16, keeping all other training hyperparameters identical to those used for PubLayNet. This consistency ensures a fair comparison across datasets.

\begin{table*}[t]
  \caption{Document Layout Detection mAP @ IoU [0.50:0.95] on PubLayNet validation set.}
  \label{tab:publaynet}
  \centering
  \scalebox{0.8}{
  \begin{tabular}{ccccccccc}
    \toprule
    \textbf{Models} & \textbf{Year} & \textbf{\#Param} & \textbf{Text} & \textbf{Title} & \textbf{List} & \textbf{Table} & \textbf{Figure} & \textbf{mAP}\\
    \midrule
    \textbf{VSR} \citep{zhang2021vsr} & 2021 & -- & \textbf{0.967} & 0.931 & 0.947 & 0.974 & 0.964 & 0.957\\
    \textbf{TRDLU} \citep{yang2022transformer} & 2022 & -- & 0.958 & 0.921 & 0.975 & 0.976 & 0.966 & 0.959\\
    \textbf{DiT-L} \citep{li2022dit} & 2022 & 304M & 0.944 & 0.893 & 0.960 & 0.978 & 0.972 & 0.949\\
    \textbf{UDoc} \citep{gu2023unified} & 2022 & 272M & 0.939 & 0.885 & 0.937 & 0.973 & 0.964 & 0.939\\
    \textbf{VGT} \citep{da2023vision} & 2023 & 174M & 0.950 & \textbf{0.939} & 0.968 & 0.981 & 0.971 & 0.962\\
    \textbf{DETR} \citep{shehzadi2023bridging} & 2023 & -- & 0.947 & 0.918 & 0.964 & 0.981 & 0.975 & 0.957\\
    \textbf{DoPTA-HR} \citep{sr2024dopta} & 2024 & 85M & 0.944 & 0.895 & 0.957 & 0.977 & 0.970 & 0.949\\
    \midrule
    \textbf{LayoutMamba-T} & 2025 & 6.9M & 0.955 & 0.883 & 0.961 & 0.985 & 0.983 & 0.953\\
    \textbf{LayoutMamba-B} & 2025 & 22M & 0.953 & 0.902 & 0.963 & \textbf{0.986} & 0.984 & 0.957\\
    \textbf{LayoutMamba-L} & 2025 & 67M & 0.957 & 0.934 & \textbf{0.965} & 0.985 & \textbf{0.985} & \textbf{0.963}\\
    \bottomrule
  \end{tabular}
  }
\end{table*}

\begin{table*}[t]
  \caption{Document Layout Detection mAP @ IoU [0.50:0.95] on $D^4LA$ validation set.}
  \label{tab:d4la}
  \centering

  \scalebox{0.6}{
  \begin{tabular}{c|c|c|c|c|c|c|c|c|c|c}
    \toprule
    \textbf{Models} & \textbf{\#Param} & \textbf{DocTitle} & \textbf{ListText} & \textbf{LetterHead} & \textbf{Question} & \textbf{RegionList} & \textbf{TableName} & \textbf{FigureName} & \textbf{Footer} & \textbf{Number}\\
    \midrule
    \textbf{DiT-B} \citep{li2022dit} & 87M & 73.1 & 70.6 & 82.2 & 55.0 & 80.1 & 68.4 & 51.8 & 81.2 & 83.2\\
    \textbf{LayoutLMv3-Base} \citep{huang2022layoutlmv3} & 133M & 66.8 & 56.5 & 78.5 & 39.3 & 72.1 & 64.3 & 32.1 & 72.2 & 82.1\\
    \textbf{VGT} \citep{da2023vision} & 174M & 72.6 & 71.3 & 82.3 & 63.9 & 80.2 & \textbf{68.4} & 46.6 & \textbf{79.7} & \textbf{83.2}\\
    \textbf{DoPTA-HR} \citep{sr2024dopta} & 85M & 73.11 & 72.46 & 82.07 & \textbf{77.42} & 79.32 & 67.08 & \textbf{56.86} & 77.88 & 83.15\\
    \textbf{LayoutMamba-T} & 6.9M & 74.2 & 69.5 & 85.3 & 68.8 & 82.1 & 61.7 & 49.5 & 77.2 & 81.4\\
    \textbf{LayoutMamba-B} & 22M & 74.6 & 72 & 83.9 & 64.9 & 82.4 & 65.2 & 52 & 78.5 & 81.5\\
    \textbf{LayoutMamba-L} & 67M & \textbf{75.4} & \textbf{73.9} & \textbf{83.9} & 67.7 & \textbf{82.7} & 65.3 & 50 & 79.6 & 82.4\\
    \bottomrule
  \end{tabular}
  }

  \vspace{6pt}

  \scalebox{0.6}{
  \begin{tabular}{c|c|c|c|c|c|c|c|c|c|c}
    \toprule
    \textbf{Models} & \textbf{\#Param} & \textbf{ParaTitle} & \textbf{RegionTitle} & \textbf{LetterDear} & \textbf{OtherText} & \textbf{Abstract} & \textbf{Table} & \textbf{Equation} & \textbf{PageHeader} & \textbf{Catalog}\\
    \midrule
    \textbf{DiT-B} \citep{li2022dit} & 87M & 63.2 & 67.5 & 74.5 & 59.2 & 73.8 & 86.2 & 9.2 & 56.5 & 44.8\\
    \textbf{LayoutLMv3-Base} \citep{huang2022layoutlmv3} & 133M & 55.6 & 59.5 & 70.8 & 50.8 & 68.2 & 80.6 & 7.3 & 53.1 & 37.3\\
    \textbf{VGT} \citep{da2023vision} & 174M & 63.0 & 67.2 & \textbf{76.7} & 60.0 & \textbf{80.4} & 86.0 & 19.9 & 56.9 & 40.9\\
    \textbf{DoPTA-HR} \citep{sr2024dopta} & 85M & 64.07 & 65.17 & 72.7 & \textbf{61.25} & 78.25 & 86.9 & \textbf{32.26} & \textbf{58.22} & \textbf{60.98}\\
    \textbf{LayoutMamba-T} & 6.9M & 61.2 & 65.6 & 70.5 & 57.1 & 75.4 & 87.6 & 9.32 & 55.3 & 32.2\\
    \textbf{LayoutMamba-B} & 22M & 62.6 & \textbf{67.4} & 72.4 & 58.4 & 74.9 & 86.9 & 26.1 & 57.2 & 35.3\\
    \textbf{LayoutMamba-L} & 67M & \textbf{64.1} & 66.5 & 71.1 & 59.5 & 74.8 & \textbf{89.9} & 25.9 & 51.9 & 38.9\\
    \bottomrule
  \end{tabular}
  }

  \vspace{6pt}

  \scalebox{0.6}{
  \begin{tabular}{c|c|c|c|c|c|c|c|c|c|c|c}
    \toprule
    \textbf{Models} & \textbf{\#Param} & \textbf{ParaText} & \textbf{Date} & \textbf{LetterSign} & \textbf{RegionKV} & \textbf{Author} & \textbf{Figure} & \textbf{Reference} & \textbf{PageFooter} & \textbf{PageNumber} & \textbf{mAP}\\
    \midrule
    \textbf{DiT-B} \citep{li2022dit} & 87M & 86.4 & 69.7 & 71.6 & 68.8 & 66.0 & 77.2 & 83.4 & 65.5 & 58.3 & 67.7\\
    \textbf{LayoutLMv3-Base} \citep{huang2022layoutlmv3} & 133M & 81.6 & 62.5 & 60.4 & 59.4 & 59.3 & 72.2 & 74.9 & 62.1 & 52.8 & 60.5\\
    \textbf{VGT} \citep{da2023vision} & 174M & 86.2 & 71.3 & 75.5 & 70.1 & 67.6 & \textbf{76.7} & 85.6 & \textbf{66.5} & 58.7 & 68.8\\
    \textbf{DoPTA-HR} \citep{sr2024dopta} & 85M & 85.75 & \textbf{71.4} & \textbf{76.31} & 70.3 & \textbf{70.66} & 75.73 & 84.45 & 65.82 & \textbf{60.64} & \textbf{70.72}\\
    \textbf{LayoutMamba-T} & 6.9M & 87.8 & 64.7 & 71.3 & 68.2 & 63.5 & 75.4 & 85.6 & 61.3 & 48.7 & 66.3\\
    \textbf{LayoutMamba-B} & 22M & 88.5 & 68.5 & 70.6 & 70.7 & 64.1 & 76.1 & 85 & 62.1 & 52.1 & 67.9\\
    \textbf{LayoutMamba-L} & 67M & \textbf{89.1} & 71.3 & 72.2 & \textbf{71.2} & 65.6 & 76.4 & \textbf{86.7} & 62.9 & 54.6 & 69.4\\
    \bottomrule
  \end{tabular}
  }
\end{table*}

\begin{table}[t]
  \centering
  \caption{Document Layout Detection mAP @ IoU [0.50:0.95] on $D^4LA$ validation set with different GOPLoss weight.}
  \label{tab:GOPLossTable}
  \scalebox{0.75}{
  \begin{tabular}{c|ccccc}
    \toprule
    \textbf{GOPLoss Weight} & \textbf{0.0} & \textbf{1.0} & \textbf{3.0} & \textbf{5.0} & \textbf{7.5}\\
    \midrule
    \textbf{mAP} & 57.9 & 58.2 & 58.2 & \textbf{58.4} & 57.6\\
    \bottomrule
  \end{tabular}
  }
\end{table}

\begin{table}[t]
  \centering
  \caption{Document Layout Detection mAP @ IoU [0.50:0.95] on $D^4LA$ validation set.}
  \label{tab:ablation}
  \scalebox{0.65}{
  \begin{tabular}{c|c|c|c}
    \toprule
    \textbf{VRSSBlock} & \textbf{Attention} & \textbf{\#Param} & \textbf{mAP}\\
    \midrule
    & & 5.8 & 60.7\\
    \midrule
    \checkmark & & 6.2 & 61.2\\
    \midrule
    \checkmark & \checkmark & 6.9 & 63.3\\
    \bottomrule
  \end{tabular}
  }
\end{table}

\subsection{Comparison to Existing Approaches}

We conducted extensive experiments on PubLayNet, comparing LayoutMamba with state-of-the-art vision Transformer backbones for document layout analysis, as summarized in Table~\ref{tab:publaynet}, where LayoutMamba achieves significant improvements, particularly in challenging categories such as ``list,'' ``table,'' and ``figure.'' LayoutMamba-T attains a comparable mAP of 95.1 to the larger DoPTA-HR model while reducing parameters from 85M to 6.9M, and LayoutMamba-L achieves a state-of-the-art mAP of 96.3, highlighting the effectiveness of our hybrid architecture. On the more challenging $D^4LA$ dataset (Table~\ref{tab:d4la}), characterized by diverse real-world scenarios, finer-grained categories, and data imbalance leading to lower mAP scores, LayoutMamba shows strong resilience; LayoutMamba-T outperforms LayoutLMv3-Base by 5.8\% mAP despite the latter having nearly $20\times$ more parameters, and achieves state-of-the-art results in key categories including DocTitle (88.2\% AP), Paratext (85.1\% AP), and RegionKV (84.5\% AP).

\subsection{Ablation Studies}

To rigorously validate the architectural efficacy and design rationale of our proposed framework, we conduct a series of systematic ablation experiments on the challenging document layout detection task, utilizing the $D^4LA$ validation set. Our investigation centers on two pivotal components that underpin the superior performance of our model: (1) the evolutionary advancements in the state-space block, where our novel Vision Residual State-Space Block (VRSSBlock) extends the foundational VSSBlock design with strategic architectural refinements, and (2) the synergistic integration of multi-head attention, which further amplifies the model's contextual reasoning capabilities.

Our reference baseline employs the original VSSBlock paradigm, leveraging a structured state space module for long-range dependency modeling and lightweight local convolutions for fine-grained feature extraction. As quantitatively summarized in Table~\ref{tab:ablation}, this baseline configuration comprises 5.8M parameters and achieves a mean Average Precision (mAP) of 60.7\% at stringent IoU thresholds [0.50,0.95], establishing a competitive yet improvable foundation for subsequent ablative analysis.

\paragraph{Impact of VRSSBlock Evolution.}
The VRSSBlock refines the VSSBlock architecture by integrating residual learning, gated MLP modules, and a novel SS2D operator, leading to enhanced feature transformation. More specifically, the VRSSBlock fuses a dedicated projection module, a global spatial information aggregator that leverages structured state space models, and a residual gated block for channel-wise refinement. When we replace VSSBlock with VRSSBlock (without adding multi-head attention), the parameter count increases modestly to 6.2M, but we observe a performance gain of approximately 0.5\% mAP (from 60.7\% to 61.2\%). This improvement demonstrates that the architectural evolution embedded in VRSSBlock contributes to more effective spatial and contextual feature modeling.

\paragraph{Impact of Multi-Head Attention.}
To further bolster the model's representational prowess, we strategically integrate a canonical multi-head attention module downstream of the VRSSBlock output. Here, query ($Q$), key ($K$), and value ($V$) projections are dynamically derived from the VRSSBlock-processed features, enabling adaptive recalibration across spatial and channel dimensions. This attentive contextualization augments global receptive field modeling by disambiguating salient regions via learned attention weights and harmonizing local and global feature hierarchies. The combined VRSSBlock + Attention configuration incurs a further parameter growth to 6.9M (+9.7\% vs. VRSSBlock-only), yet yields a decisive performance leap of +2.1\% mAP (from 61.2\% to 63.3\%). This super-additive gain underscores that attention synergizes with VRSSBlock's structured state space modeling, creating an irreducible performance ceiling that neither component could attain in isolation.

\begin{figure}[t]
    \centering
    \begin{minipage}[t]{0.48\columnwidth}
        \centering
        \includegraphics[width=\textwidth]{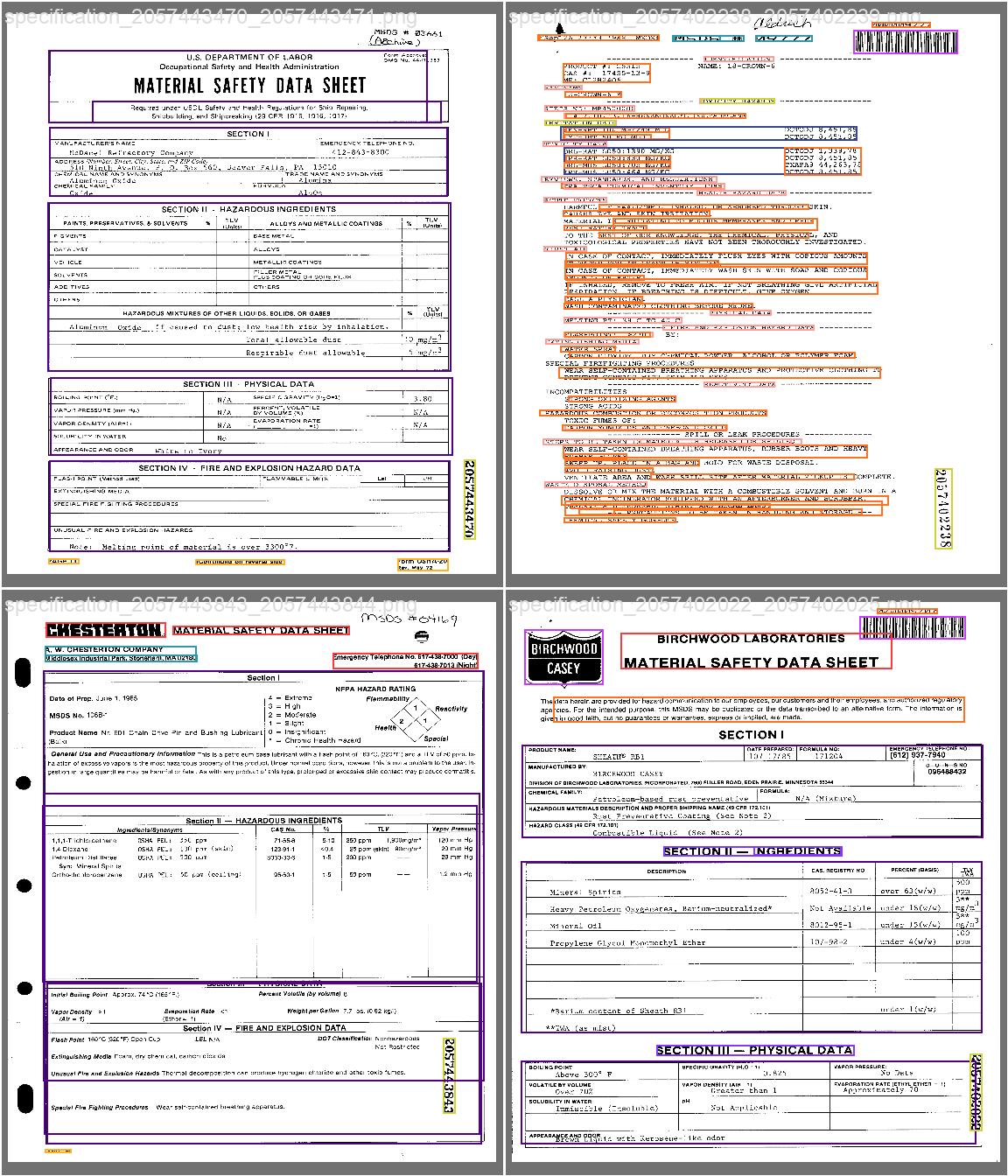}
        \vspace{-2pt}
        (a) Model with $\lambda_4=0.0$
        \label{fig:lambda_0}
    \end{minipage}%
    \hfill
    \begin{minipage}[t]{0.48\columnwidth}
        \centering
        \includegraphics[width=\textwidth]{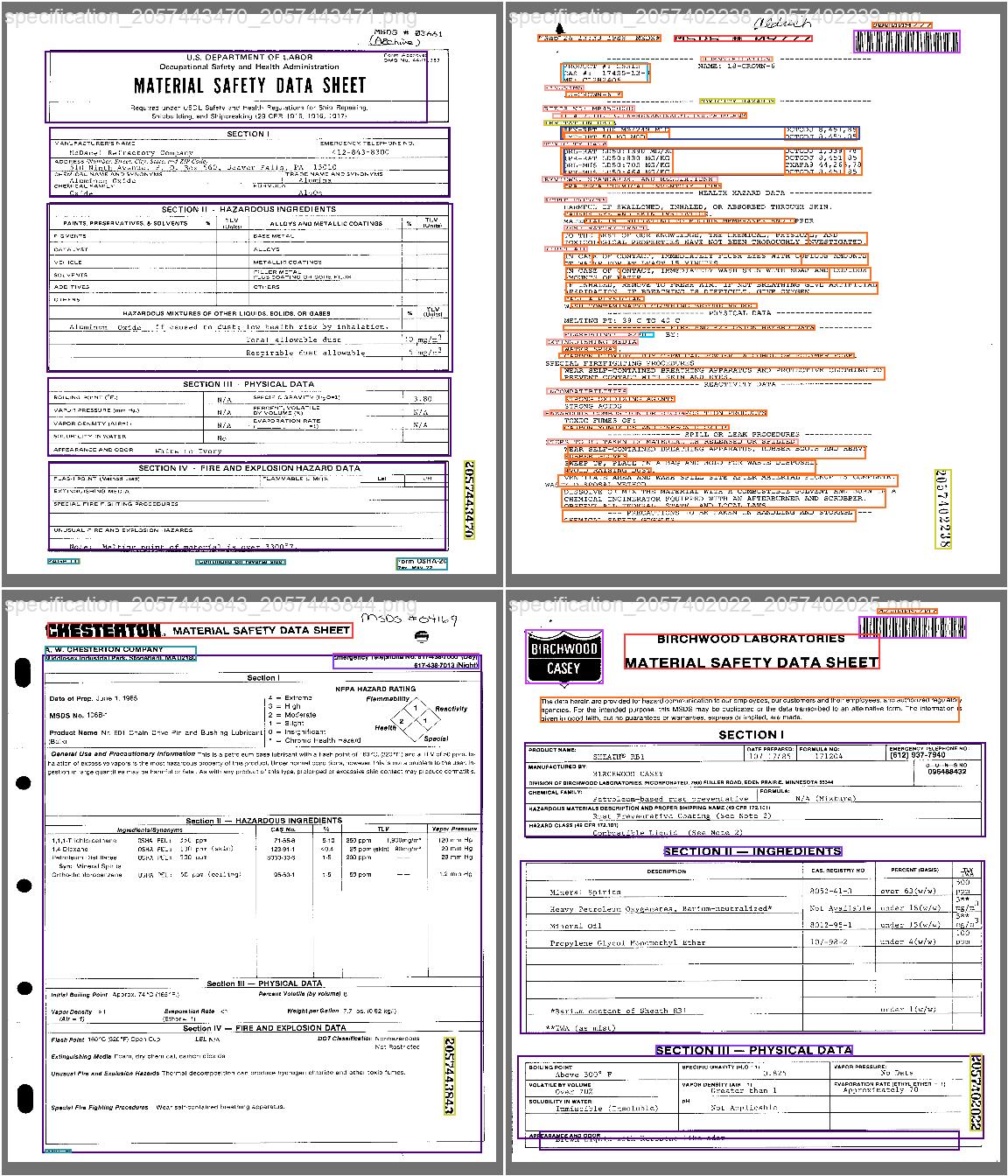}
        \vspace{-2pt}
        (b) Model with $\lambda_4=5.0$
        \label{fig:lambda_5}
    \end{minipage}
    \caption{Visual comparison of detection results under different GOPLoss weights.}
    \label{fig:GOPLoss}
\end{figure}

\paragraph{Geometric Overlap Penalty Loss.}
To systematically evaluate the impact of GOPLoss weight on the performance of the document layout analysis model, we designed a set of experiments on the $D^4LA$ benchmark dataset. Based on the trade-off between training efficiency and computational cost, the experiment set the training period to 100 epochs. We selected GOPLoss weights $\lambda_4\in\{0.0,1.0,3.0,5.0,7.5\}$ for a systematic ablation study, aiming to explore the optimal weight configuration and its regulatory mechanism on the detection performance of different layout categories.

The experimental results are shown in Table~\ref{tab:GOPLossTable}. They indicate that GOPLoss can effectively enhance the model's ability to model complex layouts through geometric constraint optimization. However, fine-tuned weight adjustment is required to achieve a multi-objective optimization balance. When $\lambda_4=5.0$, the model achieves the optimal overall performance (mAP = 58.4), which is 0.5 percentage points higher than the baseline model ($\lambda_4=0.0$, mAP = 57.9), verifying the promoting effect of moderate geometric constraints on global detection performance. However, when $\lambda_4$ increases to 7.5, the mAP significantly drops to 57.6. It is speculated that excessive geometric regularization causes the model to deviate from the optimization direction of the basic detection task. This phenomenon reveals that there is a critical threshold for GOPLoss weight. Beyond this threshold, competitive suppression will occur between geometric constraints and semantic feature learning.

To delve deeper into the optimization mechanism of the GOPLoss, a visual comparative analysis was conducted on the models with $\lambda_4=0.0$ and $\lambda_4=5.0$ (as shown in Fig.~\ref{fig:GOPLoss}). The experimental results revealed that upon the introduction of GOPLoss, the overlap of target boxes within dense text regions diminished. This phenomenon suggests that geometric constraints effectively mitigated the redundant prediction issue of adjacent detection boxes. Concurrently, the number of detection boxes generated by the model decreased, which is advantageous for enhancing the inference speed. This outcome can be ascribed to the model's adaptive suppression of low-confidence redundant predictions.

\section{Conclusion}

We propose a hybrid architecture combining Vision Transformers and State Space Models (Mamba) for Document Layout Analysis (DLA). Our approach uses sparse attention for local feature extraction and Mamba's linear complexity for global context modeling, achieving state-of-the-art results on PubLayNet and $D^4LA$ benchmarks with $8\times$ fewer parameters. We introduce a Geometric Overlap Penalty Loss to address low-overlap bias in layout predictions. Our method demonstrates Mamba's potential for efficient multimodal document understanding.

\bibliographystyle{arxiv}
\bibliography{ref}

\end{document}